\documentclass[10pt,twocolumn,letterpaper]{article}

\usepackage{iccv}
\usepackage{times}
\usepackage{epsfig}
\usepackage{graphicx}
\usepackage{amsmath}
\usepackage{amssymb}

\usepackage[numbers]{natbib} 
\usepackage{diagbox}
\usepackage{makecell}
\usepackage{multirow}
\usepackage{booktabs} 
\usepackage{rotating} 

\usepackage{subcaption}

\DeclareMathOperator*{\argmin}{argmin}
\usepackage{bm}
\newcommand*{\mat}[1]{\bm{#1}}
\DeclareMathOperator{\mydist}{d}

\DeclareMathOperator{\vect}{vec}
\usepackage[pagebackref=true,breaklinks=true,letterpaper=true,colorlinks,bookmarks=false]{hyperref}

\iccvfinalcopy 

\def\iccvPaperID{8} 

\ificcvfinal\fi

\usepackage{xspace}
\def\eg{\emph{e.g.}\xspace}
\def\ie{\emph{i.e.}\xspace}

\usepackage[utf8]{inputenc}

\title{Symmetry Aware Evaluation of 3D Object Detection and Pose Estimation in Scenes of Many Parts in Bulk}
\author{Romain Brégier$^{1, 2}$, Frédéric Devernay$^2$, Laetitia Leyrit$^1$, James L. Crowley$^{2}$\\
$^1$ Siléane, Saint-Étienne\\
$^2$ Univ. Grenoble Alpes, Inria, CNRS, Grenoble INP, LIG, F-38000 Grenoble\\
{\tt\small \{r\}.\{bregier\} at sileane.com} -- {\tt\small \{frederic\}.\{devernay\} at inria.com}}

\begin{document}
\maketitle

\begin{abstract}

While 3D object detection and pose estimation has been studied for a long time, its evaluation is not yet completely satisfactory. Indeed, existing datasets typically consist in numerous acquisitions of only a few scenes because of the tediousness of pose annotation, and existing evaluation protocols cannot handle properly objects with symmetries.
This work aims at addressing those two points. We first present automatic techniques to produce fully annotated RGBD data of many object instances in arbitrary poses, with which we produce a dataset of thousands of independent scenes of bulk parts composed of both real and synthetic images. We then propose a consistent evaluation methodology suitable for any rigid object, regardless of its symmetries.
We illustrate it with two reference object detection and pose estimation methods on different objects, and show that incorporating symmetry considerations into pose estimation methods themselves can lead to significant performance gains. The proposed dataset is available at \url{http://rbregier.github.io/dataset2017}.

\end{abstract}


\section{Introduction}
Detecting instances of 3D rigid objects and estimating their poses given visual data is of major interest for applications such as augmented reality, scene understanding and robotics, and has been an open field of research since the early days of computer vision. Nonetheless and despite important progresses in this field, existing 3D pose estimation methods still fundamentally cannot deal with any kind of rigid object, since they rely on the assumption that the pose of a rigid object corresponds to a single 6-degrees-of-freedom rigid transformation~\citep{johnson1999, mian2006, drost2010, hinterstoisser2013, tejani2014, hodan2016}. This hypothesis leads to ambiguities when dealing with objects showing some proper symmetries -- \ie invariances under some rigid transformations -- notably for evaluation purposes, as the pose of such an object can be represented by multiple rigid transformations. Symmetries are actually common among manufactured objects, and while ad hoc solutions to this issue have been proposed for performance evaluation~\citep{hinterstoisser2013, hodan2016}, they consist in relaxing the pose estimation validation criteria and are therefore not suited for applications requiring precise positioning. 

On an other level, a scenario of particular practical interest for object detection and pose estimation consists in the so-called \emph{bin-picking} problem, where instances of a rigid object have to be detected and localized within a bin containing many instances in bulk.
Despite the active research on 3D pose estimation, the current state of the art for bin-picking is relatively unknown. Secrecy is indeed the norm regarding the performances of industrial solutions, and available object recognition datasets are not representative of this scenario, as they typically consider only a limited number of object instances lying on a flat surface on a given face, because of the cost of pose annotation.

This article tries to address those issues.
In section~\ref{sec:generating_datasets}, we propose automatic techniques for generating fully annotated real range image and synthetic RGBD datasets of many instances of objects in arbitrary poses, with no redundancy between the acquisitions, contrary to existing approaches.
We suggest in section~\ref{sec:dealing_with_any_object} to take the object symmetries into account for both evaluating accuracy and improving performances of existing pose estimation methods.
Section~\ref{sec:evaluation_methodology} is focused on the evaluation problem, for which we formalize performance metrics in order to better deal with scenes containing many object instances. 
These developments are supported by experiments on two well-established object detection and pose estimation methods, in section~\ref{sec:experiments}.

\section{Related work on 3D object pose estimation evaluation}
\label{sec:related_work}
Several approaches  have been proposed to evaluate 3D object detection and pose estimation for manipulation tasks based on robotic experiments~\citep{horn1983, liu2012, buchholz2010, rodrigues2012}, but pose accuracy is difficult to evaluate in such scenario due to the lack of ground truth, and its online nature makes reproducibility difficult.
The use of datasets annotated with ground truth poses of every visible object instance instead enables a more quantitative and reproducible offline evaluation. Over the years, several publicly available datasets emerged and we summarize the characteristics of major ones in table~\ref{tab:datasets}. Other 3D object recognition datasets of interest obviously exist (see notably \citet{firman2016}) but we focus here on the case of rigid objects without intra-class variability.
\begin{table*}
\centering
\caption{\label{tab:datasets}Characteristics of typical rigid object detection and pose estimation datasets.}
\footnotesize
We distinguish between \emph{localization} and \emph{detection}~\citep{hodan2016}, depending on whether the number of instances to retrieve is constant or not. A dataset is considered \emph{redundant} if it contains numerous acquisitions of the same scene.\\
\begin{tabular}{|c|c|c|c|c|c|c|c|}
\hline
Dataset & Modality &  \parbox[c]{1.5cm}{\centering Problem class} & \parbox[c]{1.3cm}{\centering Multiple \\ instances} & \parbox[c]{1.3cm}{\centering Multiple \\ objects} &  \parbox[c]{2cm}{\centering Absence of data \\ redundancy} & \parbox[c]{1.3cm}{\centering Pose \\ variability} & Clutter \\
\hline
\citet{mian2006} &  Point cloud & Localization & no & yes & yes & limited & no \\
\citet{aldoma2012} & Colored point cloud & Detection & no & yes & partial & limited & no \\
Hinterstoisser \etal~\citep{hinterstoisser2013, brachmann2014} & RGBD & Localization & no & yes & no & limited & yes \\
Desk3D~\citep{bonde2014} & Point cloud & Detection & no & yes & no & limited & yes\\ 
\citet{tejani2014} & RGBD & Localization & yes & no & no & limited & yes \\
\citet{doumanoglou2016} & RGBD & Detection & yes & yes & no & high & limited \\
T-LESS~\citep{hodan2017} & RGBD & Detection & yes & yes & no &  high & limited\\
\hline
\end{tabular}
\end{table*}
Those datasets (except for T-LESS~\citep{hodan2017}) consist in views of scenes of a few objects lying on a given face on a table, acquired from pan-tilt viewpoints with only limited roll along the sensor axis. As such, they provide useful material for working on tasks such as indoor scene understanding, but their limited variability of poses relative to the camera is not representative of the problem of localizing objects in arbitrary poses.

\begin{figure}
\center
\small
\newcommand{\myfigdatasetwidth}{2cm}
\includegraphics[width = \myfigdatasetwidth]{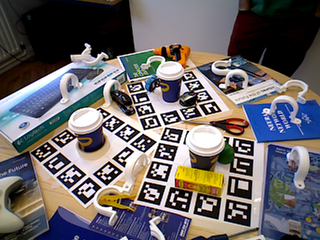}
\includegraphics[width = \myfigdatasetwidth]{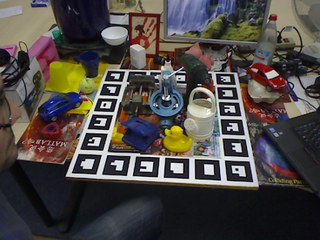}
\includegraphics[width = \myfigdatasetwidth]{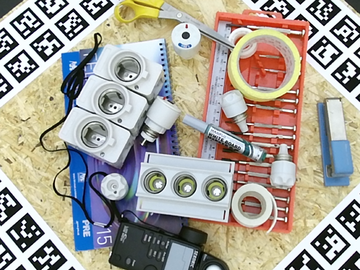}
\includegraphics[width = \myfigdatasetwidth]{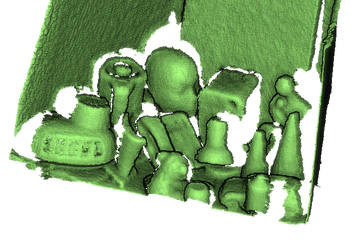}

\includegraphics[width = \myfigdatasetwidth]{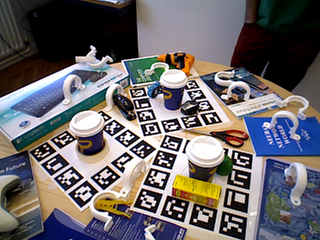}
\includegraphics[width = \myfigdatasetwidth]{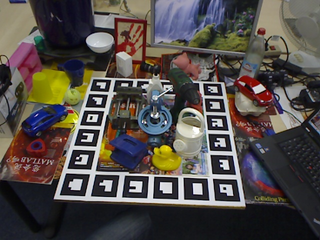}
\includegraphics[width = \myfigdatasetwidth]{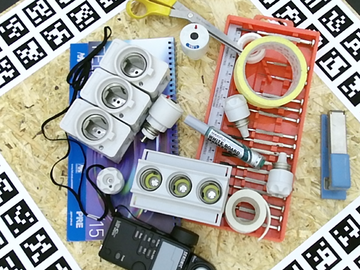}
\includegraphics[width = \myfigdatasetwidth]{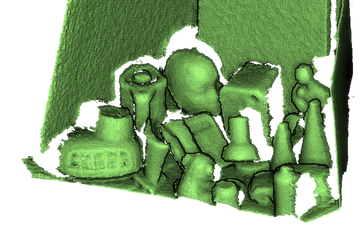}
\\
\caption{\label{fig:dataset_redundancy}Samples exhibiting the strong correlation of data within existing datasets. From left to right: datasets of Tejani \etal~\citep{tejani2014}, Hinterstoisser \etal~\citep{hinterstoisser2013, brachmann2014}, T-LESS~\citep{hodan2017} and Desk3D~\citep{bonde2014}.}
\end{figure}

Moreover, and because manually annotating objects poses in each image is a tedious process, datasets of more than a few tens of images~\citep{hinterstoisser2013, tejani2014, bonde2014, hodan2017, doumanoglou2016} rely on automatic annotation techniques. Objects instances are typically placed at given poses relatively to some fiducial markers, whose automatic detection in the scene enable to compute the poses of instances relatively to the camera. This approach enables to easily generate large annotated datasets, however consisting in numerous acquisitions of a few scenes from different viewpoints (see figure~\ref{fig:dataset_redundancy}).
This strong correlation between data samples makes those datasets less representative of the data distribution in a genuine application, and training or performance evaluation on such datasets is therefore likely to suffer from some overfitting effects.

\paragraph{Pose accuracy} An essential part of the evaluation methodology consists in selecting a criterion to decide whether or not a pose hypothesis matches a ground truth pose. Such matching criterion is typically defined based on a measure of similarity between the pose hypothesis and the ground truth. Several similarity measures have been considered for this task, such as the \emph{Intersection over Union} ratio of 2D silhouettes, the translational and rotational errors~\citep{drost2010}, and the average displacement of vertices of a 3D model of the object~\citep{hinterstoisser2013}. The first criterion, while widely used for 2D object localization, is however not suited for 3D pose estimation, as multiple poses may have similar projected silhouette on the image plane. The latter two are not well defined for objects with symmetry properties, since there is no unique displacement between two poses of a symmetric object.
The most widespread workaround to this issue consists in the use of an \emph{ad hoc} dissimilarity measure~\citep{hinterstoisser2013}, based on the distance between vertices of a 3D model  $\mathcal{M}$ of the object at those given poses
\begin{equation}
\label{eq:distance_hinterstoisser}
\underset{\mat{x}_1 \in \mathcal{M}}{\text{avg}} \min_{\mat{x}_2 \in \mathcal{M}} \| \mat{T}_1(\mat{x}_1) - \mat{T}_2(\mat{x}_2)\|,
\end{equation}
where $\mat{T}_1, \mat{T}_2 \in SE(3)$ are active rigid transforms representing the poses.
However, such criterion remains problematic, as it cannot distinguish between poses of similar 3D shapes, such as the flipped poses of coffee cups depicted figure~\ref{subfig:flaw_hinterstoisser_distance}.
\citet{hodan2016} recently suggested the use of an \emph{ambiguity-invariant pose error function}, considering a pose hypothesis as valid if and only if it is plausible given the available data.
While not denying the interest of dealing with ambiguities, especially in the context of active vision, we consider this approach problematic for evaluation, since numerous applications rely on a precise pose estimation and cannot be satisfied by plausible hypotheses. This is all the more true when dealing with highly occluded objects, for which the number of plausible hypotheses can be infinitely large.

\paragraph{Precision and recall}
Once able to classify pose results, one can define metrics to quantify the performances of a given algorithm. In the case of pose estimation of a single object instance per scene, performances are typically described by the recognition rate~\citep{johnson1999, drost2010, hinterstoisser2013}, that is the fraction of scenes for which the pose returned is the correct one on the whole dataset.
This metric is however insufficient for scenes containing an unknown number of instances, as it provides no information on the false positive results produced. In such case, the performance is typically described in terms of precision and recall~\citep{tejani2014}, averaged over all the data samples. 
Such approach however focuses on the exhaustive retrieval of every instance in each scene, which might not be the actual objective of use cases dealing with scenes of many instances.

In this article, we propose a methodology to generate both real and synthetic data of many object instances in arbitrary poses, with no dependency between data samples, in order to overcome the limitations of existing datasets, and metrics adapted for performance evaluation in scenes of many object instances, with potential symmetries.


\section{Generating annotated datasets of many instances}
\label{sec:generating_datasets}
We propose in this section two methods to generate datasets of independent scenes of many object instances in arbitrary poses, automatically annotated with the poses of those objects. The first method is based on the annotation of real range data thanks to the use of tagged object instances. The second one relies on computer simulations. We consider in our experiments multiple instances of a single object because it is representative of the bin-picking problem, however multiple objects could be considered in a similar fashion.

\subsection{Real data}
\label{subsec:real_data}
\begin{figure}
\small
\centering
\begingroup%
  \makeatletter%
  \providecommand\color[2][]{%
    \errmessage{(Inkscape) Color is used for the text in Inkscape, but the package 'color.sty' is not loaded}%
    \renewcommand\color[2][]{}%
  }%
  \providecommand\transparent[1]{%
    \errmessage{(Inkscape) Transparency is used (non-zero) for the text in Inkscape, but the package 'transparent.sty' is not loaded}%
    \renewcommand\transparent[1]{}%
  }%
  \providecommand\rotatebox[2]{#2}%
  \ifx\svgwidth\undefined%
    \setlength{\unitlength}{230bp}%
    \ifx\svgscale\undefined%
      \relax%
    \else%
      \setlength{\unitlength}{\unitlength * \real{\svgscale}}%
    \fi%
  \else%
    \setlength{\unitlength}{\svgwidth}%
  \fi%
  \global\let\svgwidth\undefined%
  \global\let\svgscale\undefined%
  \makeatother%
  \begin{picture}(1,0.66521739)%
    \put(0,0){\includegraphics[width=\unitlength]{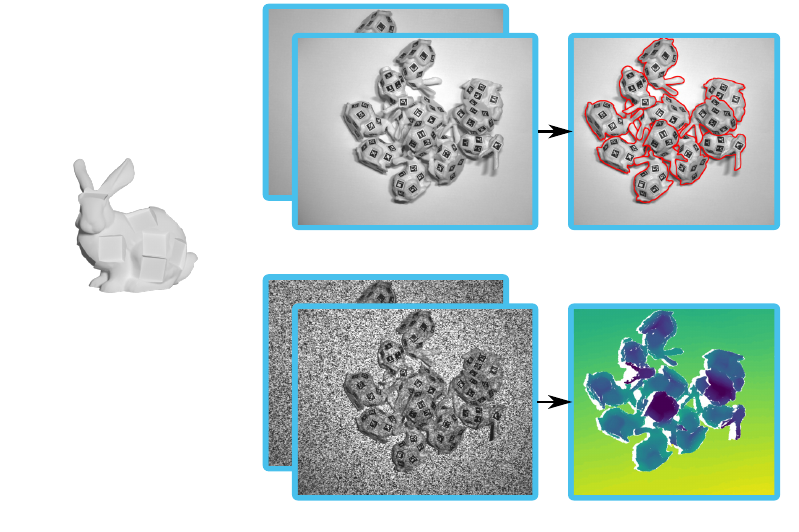}}%
    \put(0.50168878,0.3414861){\color[rgb]{0,0,0}\makebox(0,0)[b]{\smash{Uniform lighting}}}%
    \put(0.84445877,0.3414861){\color[rgb]{0,0,0}\makebox(0,0)[b]{\smash{Annotated poses}}}%
    \put(0.16230383,0.18835591){\color[rgb]{0,0,0}\makebox(0,0)[b]{\smash{\parbox[b][][c]{2cm}{\centering 3D model with markers sites}}}}%
    \put(0.50137224,0.00273313){\color[rgb]{0,0,0}\makebox(0,0)[b]{\smash{Textured lighting}}}%
    \put(0.84431267,0.00273313){\color[rgb]{0,0,0}\makebox(0,0)[b]{\smash{Depth map}}}%
  \end{picture}%
\endgroup%

\caption{\label{fig:experimental_setup}Automatic production of annotated range images of many object instances in bulk. Range images are produced by active stereo matching (left), while ground truth poses are annotated by detecting, in intensity images, markers placed on object instances (right).}
\end{figure}
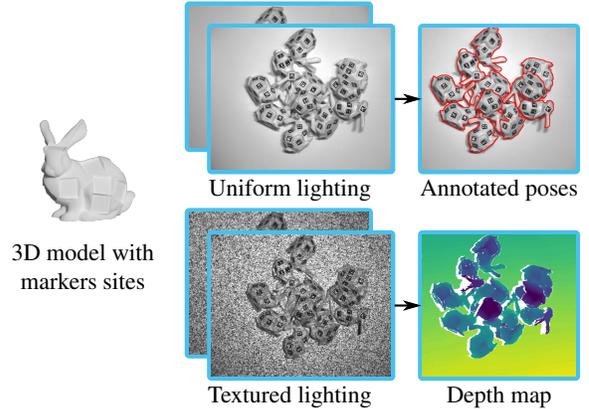

We cover the surface of each instance of object with a set of unique fiducial markers~\citep{garrido-jurado2014}, densely enough such that at least one marker is always visible from any point of view.
The pose of a marker relative to its corresponding object instance is assumed to be known, and in our experiments, we ensured the precise positioning of markers by carving their locations on the 3D model of the object, and by producing its instances by 3D printing.
This shape modification is not required but is merely performed here for convenience, and precise positioning of markers could be achieved \eg through the use of a specific assembly jig.
Using a binocular stereoscopic system, we produce a range image for each scene through the use of a pseudo-random pattern projector and an off-the-shelf stereo matching algorithm.
We also acquire intensity images from both camera with a diffuse lighting, depicted in figure~\ref{fig:experimental_setup}. This second modality is exploited in order to automatically recover the ground truth poses.

Visible markers can indeed be quite reliably detected, identified and localized within the intensity images. Because each marker is uniquely assigned to a given instance of object, object detection is straightforward: a given instance is considered present in the scene if and only if at least one of its marker is detected.
The potential occlusion of every marker of an instance is not a major issue for this approach, as it only occurs when the instance is nearly entirely occluded -- provided that the object is densely enough covered by small enough markers -- and pose annotation is of limited interest such case.

For each detected instance, marker detection in intensity image $j \in \lbrace 1, 2 \rbrace$ provides us with the 2D corners coordinates $\mat{p}_{i,j} \in \mathbb{R}^2, i \in M_j$ of the detected markers associated with the instance. Because the pose of those markers relative to the object instance is known, each $\mat{p}_{i,j}$ can be put in relation with its corresponding 3D point $\mat{X}_{i,j} \in \mathbb{R}^3$ within the object frame. The pose $[\hat{\mat{R}},\hat{\mat{t}}]$ of the considered instance -- described by a $3 \times 3$ rotation matrix $\hat{\mat{R}}$ and a 3D translation vector $\hat{\mat{t}}$ -- can therefore be estimated by solving the multiview perspective-n-point problem
\begin{equation}
[\hat{\mat{R}},\hat{\mat{t}}] = \argmin_{\mat{R},\mat{t}} \sum_{j \in \lbrace 1, 2 \rbrace} \sum_{i \in M_j} \| \pi_j(\mat{R} X_{i,j} + \mat{t}) - \mat{p}_{i,j} \|^2,
\end{equation}
where $\pi_j$ represents the projection from a 3D point in the reference coordinate system onto the image plane of camera $j$. 
The annotation procedure is therefore automatic, but we proceed in our experiments to a visual validation stage to ensure the quality of the ground truth. Validation of about 700 scenes required less than 40min for a single person, and 6\% of images were discarded due to failures of the marker detector used, consisting either in false positives, false negatives, mislabeled markers or imprecision in the localization of markers' corners. This failure rate could probably be diminished significantly through more robust markers detection, but we considered it acceptable given that scene acquisition and validation of automatic annotations are quite fast to perform.
Additional annotations such as the occlusion rate of each instance are produced by comparison with CGI-renderings of the instances at the annotated poses. 

\subsection{Synthetic data}
While effective, this marker-based approach is intrusive, in that markers remain visible in intensity images, which limit the suitability of this approach for object recognition techniques based on such modality. Those issues can be avoided through the use of synthetic data, for which ideal ground truth can be produced.
It is indeed possible to generate synthetic datasets reasonably representative of controlled environments such as an industrial setup for bin-picking, and for our experiments we produce synthetic scenes of bulk objects through the physical simulation of drop of instances into a bin, from which we synthesize top-view RGBD images (see examples figure~\ref{fig:results_samples}).

\paragraph{Sensor simulation}
Producing plausible synthetic RGB images is relatively simple thanks to existing ray-tracing renderers, but while range data simulators~\citep{gschwandtner2011} or depth noise models (\eg \citep{handa2014}) have been proposed, they remain quite sensor-dependent and their tuning is not trivial.
To overcome those issues, we render stereo image pairs of the synthetic scene lit by a virtual pseudo-random pattern projector using the Blender Cycles renderer~\citep{blender},  and perform on those images the same 3D stereo reconstruction as in our real experiments. This approach produces range images with a 3D reconstruction noise visually similar to the one observed in our real data, without the need for an explicit noise model, and despite a coarse modeling of the scene (we discuss this question further in section~\ref{subsec:comparison_real_vs_synthetic}).
The reconstruction technology we simulate here is similar to the one used in industrial cameras such as the Ensenso N10, and other triangulation-based sensors such as LASER scanners or Kinect v1 could be simulated in a similar fashion, given access to the processing algorithms used by these devices to transform raw input data into depth images.

Based on those approaches, we generated both real and synthetic datasets, considering objects with various symmetry properties, in scenes of various number of instances piled up in bulk (see fig.~\ref{fig:results_samples} and supplementary material). The \emph{clutter} dataset is an exception depicting very cluttered scenes, to illustrate the suitability of our annotation procedure in such scenario.

\section{Dealing with symmetric objects}
\label{sec:dealing_with_any_object}

Being able to quantify the accuracy of a pose estimate is a prerequisite for evaluation, but usual measures are not fit to deal with symmetric objects. In this section, we suggest the use of a distance suited to any bounded rigid object, and discuss how it can be used within 3D object detection and pose estimation techniques themselves, in order to increase performances.

\subsection{Pose distance}
\citet{bregier2016} recently proposed a pose definition valid for any rigid object -- including those with symmetries -- as a distinguishable static state of the object. Within their framework, a pose can be identified to an equivalency class $\left\lbrace \mat{T} \circ \mat{G} \middle\vert \mat{G} \in G \right\rbrace$ of $SE(3)$, defined by an active rigid transformation $\mat{T}$ and up to any rigid transformation the object is invariant to (the \emph{proper symmetry group}~$G \subset SE(3)$).

\begin{figure}
\begin{subfigure}[t]{0.49\linewidth}
\includegraphics{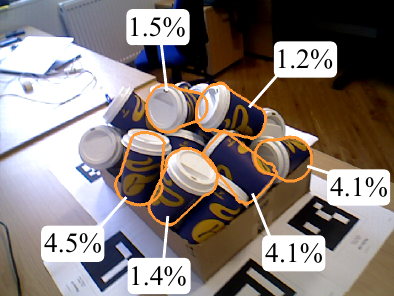}
\caption{\label{subfig:flaw_hinterstoisser_distance}\citet{hinterstoisser2013}.}
\end{subfigure}
\begin{subfigure}[t]{0.49\linewidth}
\includegraphics{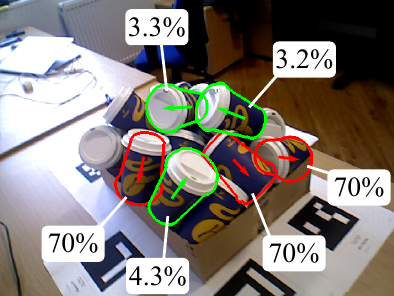}
\caption{Proposed distance.}
\end{subfigure}
\caption{\label{fig:comparison_hinterstoisser_bregier_validation_distances}Dissimilarity (in \% of the object's diameter $D$) between pose hypotheses (overlaid contours) and the corresponding ground truth poses. In this example from Doumanoglou's dataset~\citep{doumanoglou2016}, each pose hypothesis would be considered as valid according to the widespread criterion of \citet{hinterstoisser2013} for symmetric objects (dissimilarity below $10\% \cdot D$). The proposed distance on the other hand properly accounts for the object's symmetries, which enables to discriminate well true positives (green) and false positives (upside down hypotheses, in red).}
\end{figure}

\begin{figure*}
\centering
\footnotesize
\setlength\tabcolsep{1pt}
\newlength\myfigheight
\setlength\myfigheight{2.5cm}
\begin{tabular}{cccccc}
\includegraphics[height=1cm]{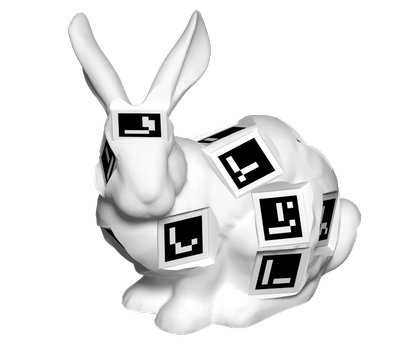}&
\includegraphics[height=1cm]{bunny.png}&
\includegraphics[height=1cm]{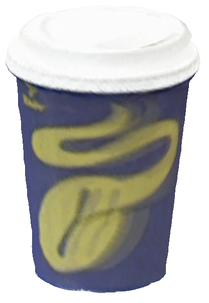}&
\includegraphics[height=1cm]{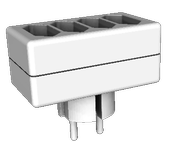}&
\includegraphics[height=1cm]{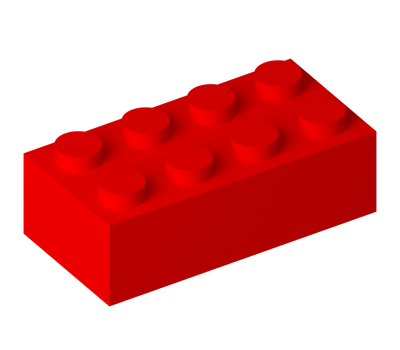}&
\includegraphics[height=1cm]{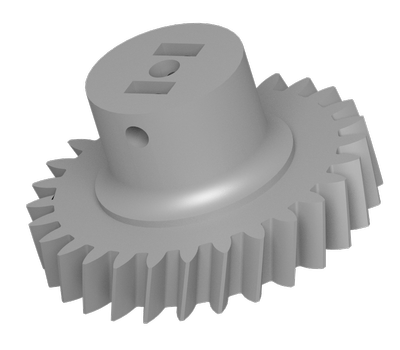}
\\
\includegraphics[height=\myfigheight]{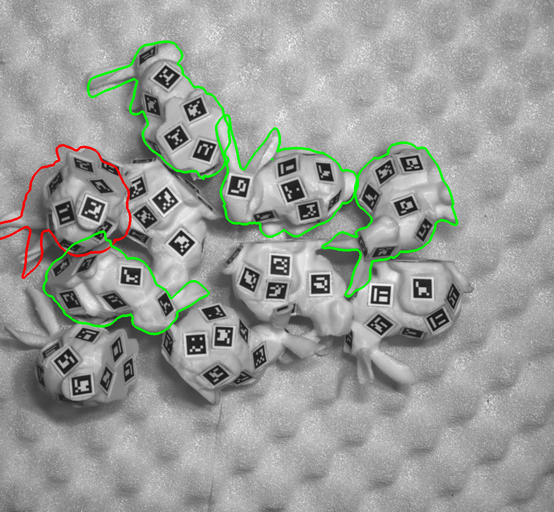}&
\includegraphics[height=\myfigheight]{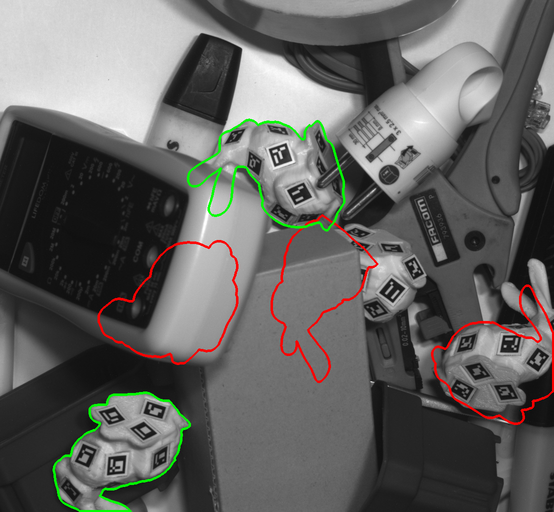}&
\includegraphics[height=\myfigheight]{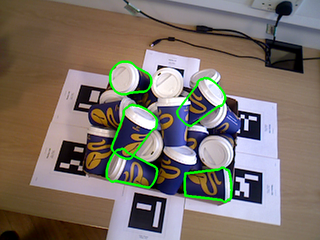}&
\includegraphics[height=\myfigheight]{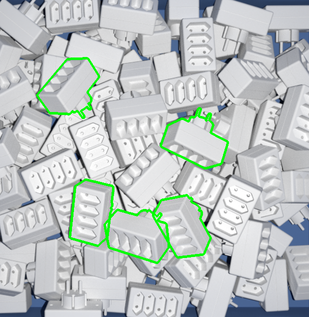}&
\includegraphics[height=\myfigheight]{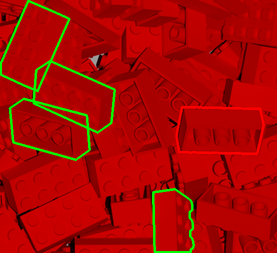}&
\includegraphics[height=\myfigheight]{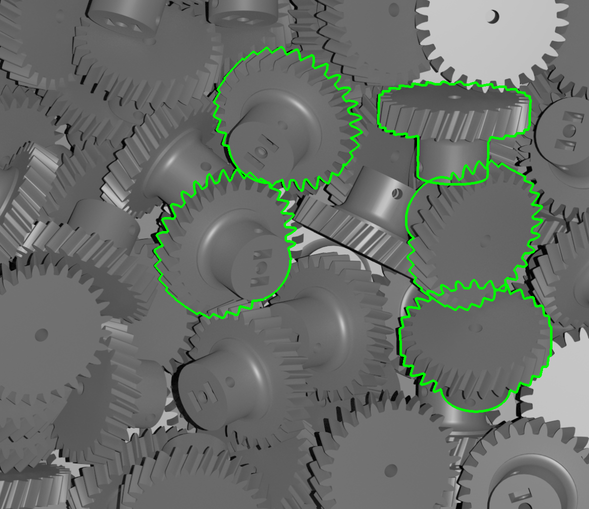}\\
\includegraphics[height=\myfigheight]{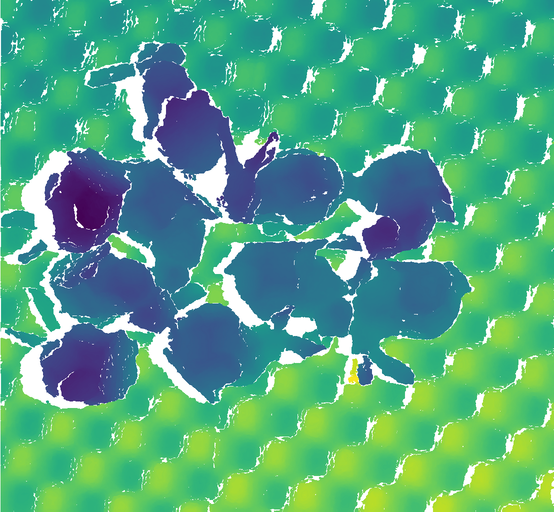}&
\includegraphics[height=\myfigheight]{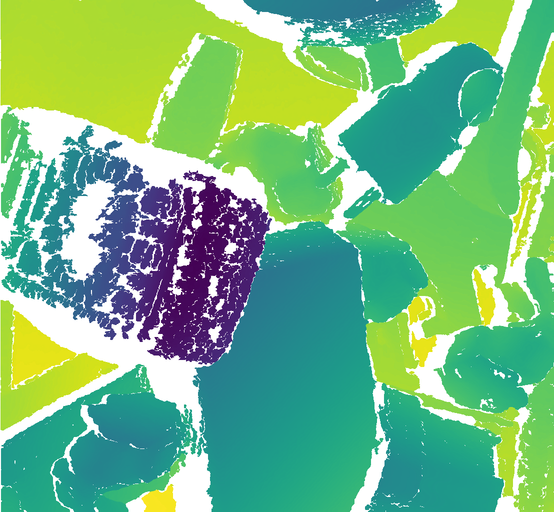}&
\includegraphics[height=\myfigheight]{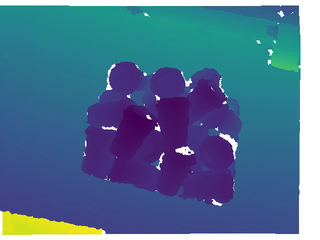}&
\includegraphics[height=\myfigheight]{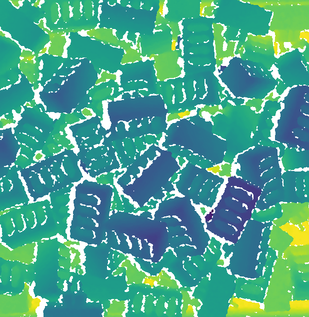}&
\includegraphics[height=\myfigheight]{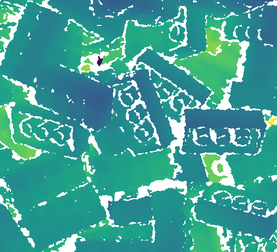}&
\includegraphics[height=\myfigheight]{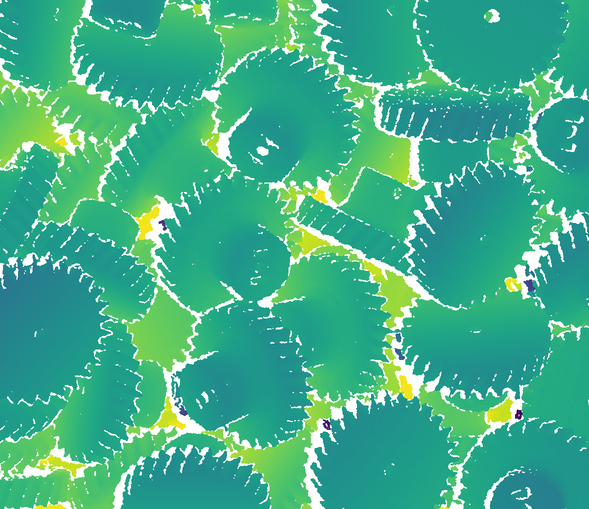}\\
markers bump&
markers clutter&
coffee cup~\citep{doumanoglou2016}&
tless 20&
brick$^*$&
gear\\[-1mm]
\multicolumn{3}{c}{\upbracefill \hspace{3pt}} & \multicolumn{3}{c}{\hspace{10pt} \upbracefill}\\
\multicolumn{3}{c}{\small Real data} & \multicolumn{3}{c}{\small Synthetic data}\\
\end{tabular}\\
\caption{\label{fig:results_samples}Examples of results from our experiments in object detection and pose estimation. Top: considered objects. Middle: RGB images, with silhouettes contours of the first 5 results returned by the PPF \emph{sym} method with post-processing, classified as true positives (green) or false positives (red). Bottom: corresponding range data. $^*$: the false positive brick highlighted in red is flipped upside down compared to the corresponding ground truth.}
\end{figure*}

Given a bounded rigid object, they propose a distance between two poses consisting in the length of the smallest displacement from one pose to an other, the length of a displacement being defined as the RMS displacement of surface points of the object
\begin{equation}
\label{eq:distance_bregier}
\begin{aligned}
&\mydist(\mathcal{P}_1, \mathcal{P}_2) \triangleq\\
& \min_{\mat{G}_1, \mat{G}_2 \in G } \sqrt{ \cfrac{1}{S} \int_{\mathcal{S}} \| \mat{T}_2 \circ \mat{G}_2 (\mat{x}) - \mat{T}_1 \circ \mat{G}_1 (\mat{x})\|^2 ds},
\end{aligned}
\end{equation}
where $S$ is the surface area of the object. This distance is physically meaningful, and accounts properly for object's symmetries contrary to the widespread measure~\eqref{eq:distance_hinterstoisser} as illustrated figure~\ref{fig:comparison_hinterstoisser_bregier_validation_distances}.
Moreover, it can be estimated efficiently in closed form, whereas the measure~\eqref{eq:distance_hinterstoisser} requires to perform systematically a computation on every vertex of the model. To this aim, the authors propose a representation of a pose $\mathcal{P}$ as a finite set of points $\mathcal{R}(\mathcal{P})$ of at most 12 dimensions, depending on the proper symmetry class of the object. Those representations are listed in table~\ref{tab:symmetries_classes} and in particular,
\begin{itemize}
\item A pose of an object without proper symmetries (such as the bunny covered with markers used in our experiments) is represented by a 12D point, consisting in the concatenation of the 3D position of its centroid $\mat{t}$ and of its rotation matrix $\mat{R}$, anisotropically scaled by a matrix $\mat{\Lambda}$ to account for the object's geometry.
\item A pose of an object with a finite non-trivial proper symmetry group is represented by several of those 12D points, two in the case of the \emph{brick} object figure~\ref{fig:results_samples} which is invariant under the group of proper symmetries $G= \left\lbrace \mat{I}, \mat{R}_z^\pi \right\rbrace$ consisting in the identity transformation and a rotation of $1/2$ turn around a given axis.
\item Poses of revolution objects (such as \emph{pepper} and \emph{gear}  figure~\ref{fig:results_samples}) are represented by 6D vectors consisting in the concatenation of the 3D position of their centroids and the direction of their revolution axis, scaled to account for the object's geometry.
\end{itemize}

Using this representation, they show that the distance between two poses $\mathcal{P}_1, \mathcal{P}_2$ can be evaluated as the Euclidean distance between any given point $\mat{p}_1 \in \mathcal{R}(\mathcal{P}_1)$ and the pointset $\mathcal{R}(\mathcal{P}_2)$:
\begin{equation}
\forall \mat{p}_1 \in \mathcal{R}(\mathcal{P}_1), \text{d}(\mathcal{P}_1, \mathcal{P}_2) = \min_{\mat{p}_2  \in \mathcal{R}(\mathcal{P}_2)} \|\mat{p}_2 - \mat{p}_1 \|.
\end{equation}
This ``nearly Euclidean'' structure enables to perform neighborhood queries such as nearest-neighbor or radius searches in an exact or approximate fashion using existing algorithms developed for Euclidean spaces (regular grids, kD-trees, \emph{etc.}). It also enables to average poses of an object with potential symmetries efficiently. Thanks to those possibilities, high-level techniques can be developed in the space of poses, such as the Mean Shift algorithm, which finds local maxima of density distributions.
\begin{table*}
\footnotesize\centering
\caption{\label{tab:symmetries_classes} Classification of every potential group of proper symmetry for a 3D bounded physical object, and expression of pose representatives enabling fast distance computations.}
\renewcommand{\arraystretch}{1.5}
\begin{tabular}{|c|c|c|c|c|}
\multicolumn{1}{c}{} &
\multicolumn{1}{c}{\includegraphics[height=1.5cm]{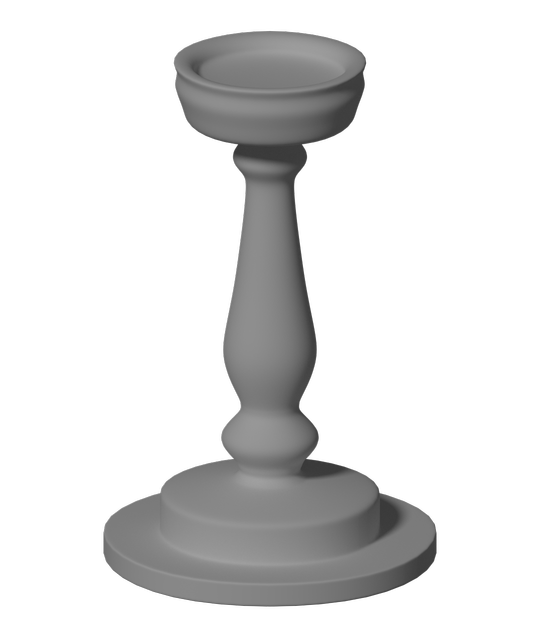}} & 
\multicolumn{1}{c}{\includegraphics[height=1.5cm]{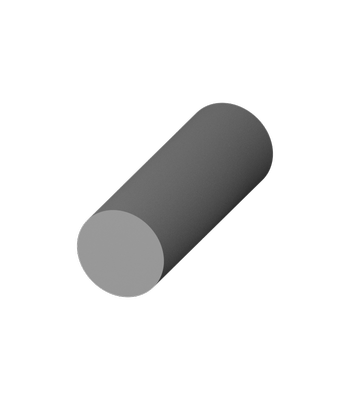}} & 
\multicolumn{1}{c}{\includegraphics[height=1.5cm]{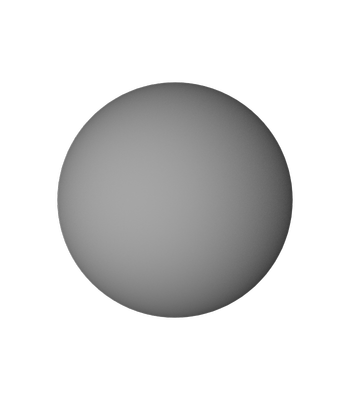}}& 
\multicolumn{1}{c}{\includegraphics[height=1.5cm]{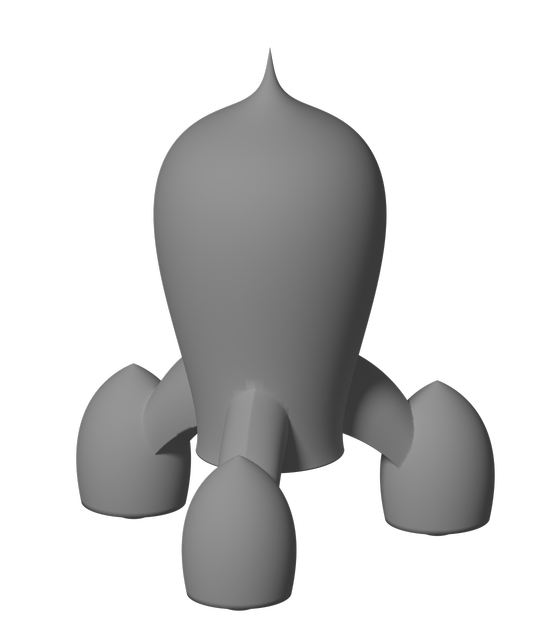}} \\
\hline
Proper symmetry class &
\makecell[c]{Revolution without \\ rotoreflection invariance} & \makecell{Revolution with \\ rotoreflection invariance} & \makecell{Spherical} & \makecell{Finite} \\
\hline
Proper symmetry group $G$ &
$\left\lbrace \mat{R}_z^\alpha \middle\vert \alpha \in \mathbb{R} \right\rbrace$ &
$\left\lbrace \mat{R}_x^\delta \mat{R}_z^\alpha \middle\vert \delta \in \left\lbrace 0, \pi \right\rbrace, \alpha \in \mathbb{R} \right\rbrace$ &
$SO(3)$ &
$G \subset SO(3)$\\
\hline
Pose representatives $\mathcal{R}(\mathcal{P})$ &
$(\lambda (\mat{R} \mat{e}_z)^\top, \mat{t}^\top)^\top \in \mathbb{R}^{6}$ &
$\left\lbrace (\pm \lambda (\mat{R} \mat{e}_z)^\top, \mat{t}^\top)^\top\right\rbrace \subset \mathbb{R}^{6}$ &
$\mat{t} \in \mathbb{R}^3$ &
\hspace{-2mm} $\left\lbrace (\vect(\mat{R} \mat{G} \mat{\Lambda})^\top, \mat{t}^\top)^\top | \mat{G} \in G \right\rbrace \subset \mathbb{R}^{12}$ \hspace{-2mm} \\
\hline
\end{tabular}\\
\textbf{Assumptions}:
Object frame $(\mat{O}, \mat{e}_x, \mat{e}_y, \mat{e}_z)$ chosen such as $\mat{O}$ to be the center of mass of the object and the $\mat{e}_z$ axis to be aligned with the symmetry axis for revolution objects.\\
\textbf{Notations}: $\mat{I}$ is the identity rotation. $\mat{R}_x^\alpha, \mat{R}_y^\alpha,\mat{R}_z^\alpha$ represent rotations of angle $\alpha \in \mathbb{R}$ around respectively $\mat{e}_x, \mat{e}_y, \mat{e}_z$ axes.\\
$\mat{\Lambda} \triangleq ( \cfrac{1}{S} \int_\mathcal{S} \mat{x} \mat{x}^\top ds )^{1/2}$,
and $\lambda \triangleq \sqrt{\lambda_r^2 + \lambda_z^2}$ for revolution objects where $\mat{\Lambda} = \text{diag}(\lambda_r, \lambda_r, \lambda_z)$.\\
\end{table*}

Given its advantages, we suggest to use this distance for quantifying the accuracy of a pose estimation. It expresses the RMS error in positioning of surface points of the object, and we define a criterion $m(p,t) \triangleq (d(p, t) < \delta)$ to decide whether or not a pose hypothesis $p$ matches a ground truth pose $t$ based on a hard threshold $\delta$, in length unit. The choice of such threshold is very application-dependent, but without further information we chose arbitrarily a value of $10\%$ of the diameter of the smallest sphere enclosing the object and centered on the centroid of object's surface.

\subsection{Considering symmetries for 3D pose estimation}
\label{subsec:considering_symmetries_for_pose_estimation}

\paragraph{Finding modes}
An important class of 3D detection and pose estimation methods are based on the aggregation of multiple votes for poses~\citep{drost2010, kim2011, rodrigues2012, tejani2014, birdal2015, kehl2015} in order to generate pose hypotheses.
They typically rely on density estimation, modes finding techniques such as Mean Shift, or clustering operations.
Those operations depend on the choice of metric over the pose space, and can actually be performed efficiently considering the distance~\eqref{eq:distance_bregier} (see~\citep{bregier2016} for details). Contrary to a metric suited for $SE(3)$, the suggested one enables to take object symmetries into account directly in the 3D pose estimation method, and to better exploit the aggregate of votes.
We test this hypothesis  in our experiments by adapting to this approach the method of~\citet{drost2010}, to which we will refer as PPF. 

\paragraph{Filtering duplicates}
More generally, considering symmetries can improve the performance of any 3D pose estimation method thanks to early duplicate removal, as one can ignore pose hypotheses too similar to an already retained one, according to the distance~\eqref{eq:distance_bregier}. Removing duplicates is a simple way to increase the precision in object detection and pose estimation, and performing it at an early stage benefits to computation time, by limiting the number of pose hypotheses to refine and validate. Equivalently given a constant number of pose hypotheses retained, duplicates removal enables to consider a larger number of truly different pose hypotheses, which should benefit to the recall.
We experiment this approach on two methods: PPF evoked previously; and the sliding window technique of~ \citet{hinterstoisser2013} (referred to as LINEMOD+).


\section{Performance metrics}
\label{sec:evaluation_methodology}
In scenes containing many parts, retrieving the pose of every single object instance is not always required. The pose of very occluded objects is indeed often ambiguous, and the retrieval of a limited number of instances is sufficient for many applications -- \eg in robotic manipulation. We propose here some adaptations of the usual performance metrics -- precision and recall -- to take those aspects into account.
We discuss the case of a scene containing potentially multiple instances of a rigid object, but multiple scenes should obviously be considered for statistical robustness through the use of aggregated metrics, such as the mean precision and recall.
The multi-object case can be handled similarly, as long as no specific relations between the objects are implied (\eg no object categories).
Let $T$ be the set of ground truth poses of object instances within the scene, and $P$ the set of result poses retrieved via an object detection and pose estimation method.

\paragraph{Instances of interest}
\label{sec:uninteresting_parts}
Only a subset $T_o \subset T$ of object instances present in the scene might be of interest to retrieve. In our experiments, we choose $T_o \triangleq \left\lbrace t \in T \middle\vert o(t) < \delta_o \right\rbrace$ as the subset of instances $t$ with an occlusion rate $o(t)$ smaller than $\delta_o = 50\%$. Given $T_o$, we define the notions of true positives ($TP$), false positives ($FP$), and false negatives ($FN$) \begin{equation}
\begin{aligned}
TP\! &=\! \left\lbrace (p,t) \in P\! \times\! T_o \middle\vert m(p, t) \land p = n_P(t) \land t = n_T(p)  \right\rbrace\\
FP\! &=\! \left\lbrace p \in P \middle\vert \lnot m(p,n_T(p)) \lor n_P(n_T(p)) \neq p \right\rbrace\\
FN\! &=\! \left\lbrace t \in T_o \middle\vert \lnot m(t, n_P(t)) \lor n_T(n_P(t)) \neq t \right\rbrace,
\end{aligned}
\end{equation}
where  $n_S(q) \triangleq \argmin_{r \in S} \mydist(q, r)$ is the nearest pose within a set $S$ from a pose $q$. Duplicates are considered as false positives with this definition, and a result corresponding to an instance whose retrieval is of no interest (within $T \setminus T_o$) is neither considered as a true nor a false positive.
The complementary notions of precision and recall can then be derived as
\begin{equation}
\label{def:precision_sensitivity}
\begin{aligned}
\text{precision} &= |TP|/(|TP| + |FP|),\\
\text{ recall} &=  |TP|/(|FN|+|TP|).
\end{aligned}
\end{equation}

\paragraph{Limited number of retrievals}
\label{sec:limited_number_of_retrievals}
To properly handle the case where the number of results $|P|$ is restricted to $n \in \mathbb{N}^*$, we propose to alter the  definition of recall as follows
\begin{equation}
\text{recall}_{\leq n \ \text{results}} = |TP| / \min(n, |FN| + |TP|).
\end{equation}

\section{Experiments}
\label{sec:experiments}

This section presents experiments performed on object detection and pose estimation using
the PPF and LINEMOD+ methods, adapted to deal with symmetric objects (see section~\ref{subsec:considering_symmetries_for_pose_estimation}). Through these, we aim to illustrate our different evaluation proposals suited for scenes depicting multiple instances of any rigid object; and also to bring out the benefits of considering symmetries for object detection and pose estimation.

\subsection{Protocol}
Our experiments are based on the LINEMOD implementation of Stefan Holtzer available in PCL~\citep{rusu2011}, and on our own implementation of PPF.
Prior to evaluation, we refine pose hypotheses thanks to a projective ICP algorithm -- an additional step often performed to improve pose estimation accuracy~\citep{hinterstoisser2013, kehl2015, kehl2016}. The evaluation therefore focuses on the ability to generate pose hypotheses within the convergence basin of poses of actual instances.
We perform experiments on both our real and synthetic datasets (some of those generated using object models from T-LESS~\citep{hodan2017}), as well as the two bin-picking datasets from~\citet{doumanoglou2016}.
Computation time is not evaluated here, as we used unoptimized code unrepresentative of the original methods.

\paragraph{Post-processing step}
The addition of a post-processing step (PP) for objects pose hypotheses has been shown to substantially improve performances~\citep{hinterstoisser2013, aldoma2016}. We evaluate this effect by considering variants of PPF and LINEMOD+ consisting in keeping the 20 best hypotheses returned by the method, scoring them according to their consistency with the input data and filtering duplicates.

\paragraph{Considering symmetries}
As discussed in section~\ref{subsec:considering_symmetries_for_pose_estimation}, we also evaluate the impact of considering the proper symmetries of the object if any (suffixed by \emph{sym} in table~\ref{tab:performances}). We use the same set of templates for LINEMOD+ in both cases, so as not to bias the comparison.


\subsection{Suitability of synthetic data for evaluation}
\label{subsec:comparison_real_vs_synthetic}
\begin{figure}
\centering
\begin{subfigure}[t]{\linewidth}
\centering
\includegraphics[width=0.4\linewidth]{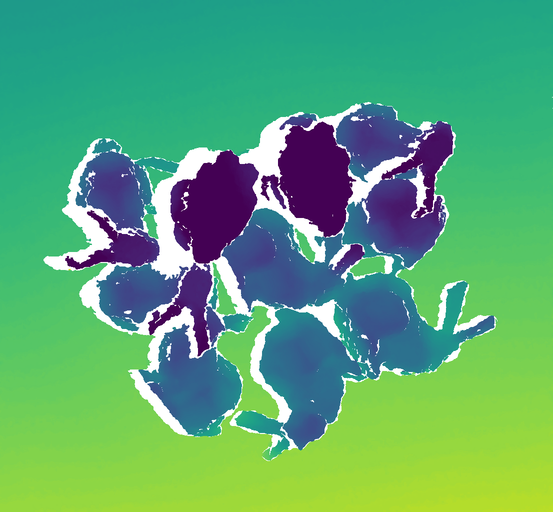}
\includegraphics[width=0.4\linewidth]{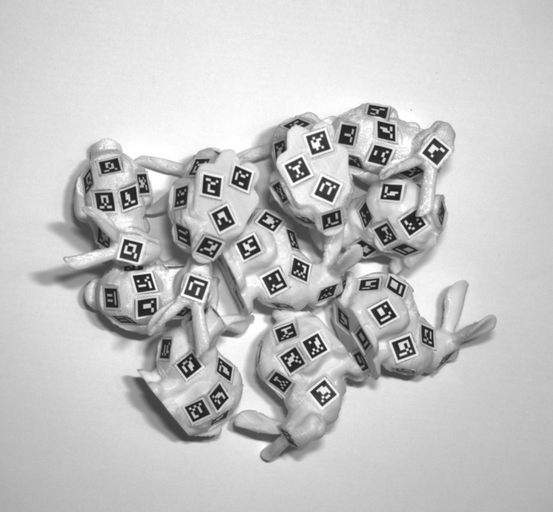}
\caption{Real data}
\end{subfigure}
\begin{subfigure}[t]{\linewidth}
\centering
\includegraphics[width=0.4\linewidth]{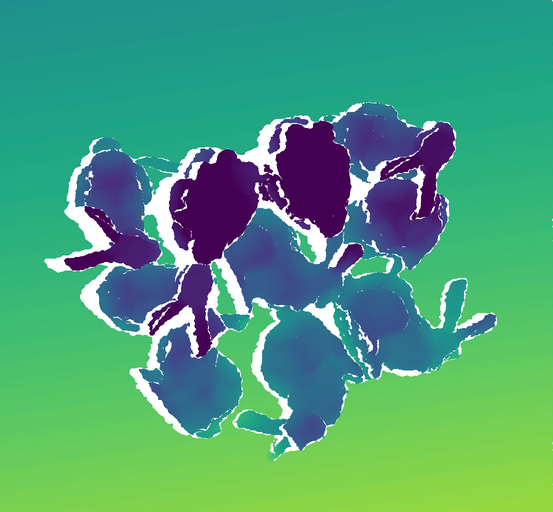}
\includegraphics[width=0.4\linewidth]{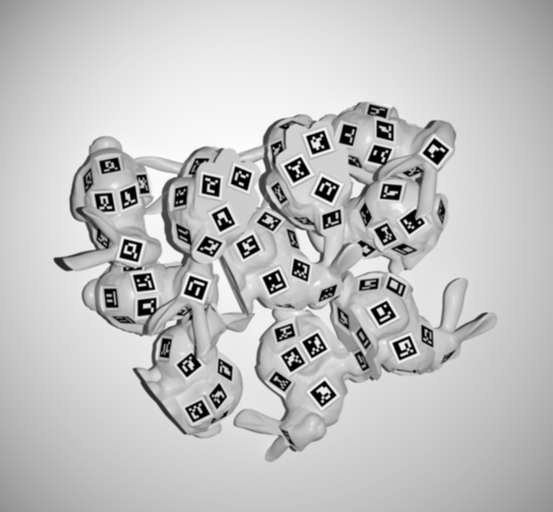}
\caption{Synthetic data}
\end{subfigure}
\caption{\label{fig:real_and_synthetic}Synthetic dataset depicting virtual copies of real scenes for comparison purposes (depth and intensity).}
\end{figure}

The use of synthetic data raises the question of its suitability for evaluation purposes. To assess the usefulness of our depth sensor simulation procedure for this task, we generate a synthetic dataset of 308 images depicting virtual copies of real scenes  of object instances lying on a flat background (\emph{markers flat}), and we perform evaluations on both datasets. 
These virtual scenes are synthesized automatically based on the pose annotations of real data produced using markers, and a plane detection algorithm for the background. Parameters of the virtual cameras used for range data generation match those of real cameras, and figure~\ref{fig:real_and_synthetic} depicts an example of synthesized data.

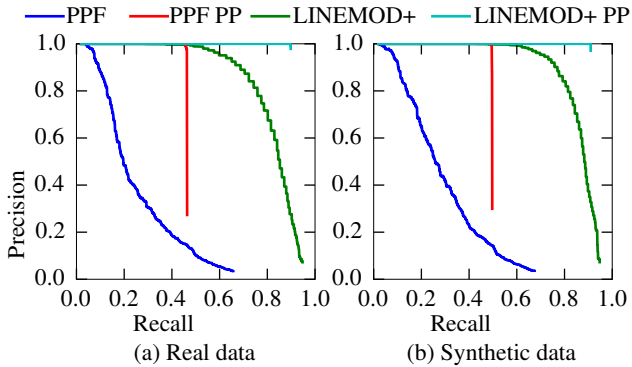
\begin{figure}
\centering
\small
\begingroup%
  \makeatletter%
  \providecommand\color[2][]{%
    \errmessage{(Inkscape) Color is used for the text in Inkscape, but the package 'color.sty' is not loaded}%
    \renewcommand\color[2][]{}%
  }%
  \providecommand\transparent[1]{%
    \errmessage{(Inkscape) Transparency is used (non-zero) for the text in Inkscape, but the package 'transparent.sty' is not loaded}%
    \renewcommand\transparent[1]{}%
  }%
  \providecommand\rotatebox[2]{#2}%
  \ifx\svgwidth\undefined%
    \setlength{\unitlength}{235.27558594bp}%
    \ifx\svgscale\undefined%
      \relax%
    \else%
      \setlength{\unitlength}{\unitlength * \real{\svgscale}}%
    \fi%
  \else%
    \setlength{\unitlength}{\svgwidth}%
  \fi%
  \global\let\svgwidth\undefined%
  \global\let\svgscale\undefined%
  \makeatother%
  \begin{picture}(1,0.57831325)%
    \put(0,0){\includegraphics[width=\unitlength]{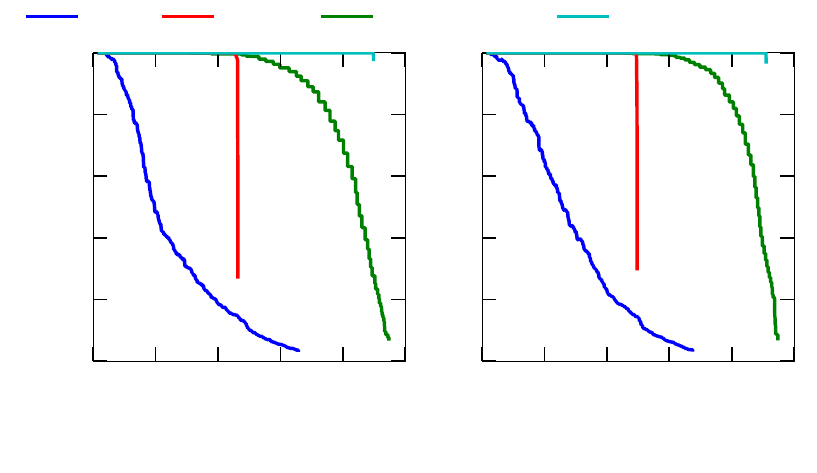}}%
    \put(0.08924541,0.09343844){\makebox(0,0)[lb]{\smash{0.0}}}%
    \put(0.16619772,0.09343844){\makebox(0,0)[lb]{\smash{0.2}}}%
    \put(0.2418218,0.09343844){\makebox(0,0)[lb]{\smash{0.4}}}%
    \put(0.31844205,0.09343844){\makebox(0,0)[lb]{\smash{0.6}}}%
    \put(0.39496269,0.09343844){\makebox(0,0)[lb]{\smash{0.8}}}%
    \put(0.47231347,0.09343844){\makebox(0,0)[lb]{\smash{1.0}}}%
    \put(0.25368687,0.05320637){\makebox(0,0)[b]{\smash{Recall}}}%
    \put(0.04743938,0.12690982){\makebox(0,0)[lb]{\smash{0.0}}}%
    \put(0.04850196,0.20239577){\makebox(0,0)[lb]{\smash{0.2}}}%
    \put(0.04690809,0.27788171){\makebox(0,0)[lb]{\smash{0.4}}}%
    \put(0.04730656,0.35336766){\makebox(0,0)[lb]{\smash{0.6}}}%
    \put(0.04750579,0.42885361){\makebox(0,0)[lb]{\smash{0.8}}}%
    \put(0.04936531,0.50433955){\makebox(0,0)[lb]{\smash{1.0}}}%
    \put(0.03042476,0.24036718){\rotatebox{90}{\makebox(0,0)[b]{\smash{Precision}}}}%
    \put(0.56528289,0.09343844){\makebox(0,0)[lb]{\smash{0.0}}}%
    \put(0.6422352,0.09343844){\makebox(0,0)[lb]{\smash{0.2}}}%
    \put(0.71785928,0.09343844){\makebox(0,0)[lb]{\smash{0.4}}}%
    \put(0.79447953,0.09343844){\makebox(0,0)[lb]{\smash{0.6}}}%
    \put(0.87100017,0.09343844){\makebox(0,0)[lb]{\smash{0.8}}}%
    \put(0.94835095,0.09343844){\makebox(0,0)[lb]{\smash{1.0}}}%
    \put(0.72972434,0.05320637){\makebox(0,0)[b]{\smash{Recall}}}%
    \put(0.52347686,0.12690982){\makebox(0,0)[lb]{\smash{0.0}}}%
    \put(0.52453944,0.20239577){\makebox(0,0)[lb]{\smash{0.2}}}%
    \put(0.52294557,0.27788171){\makebox(0,0)[lb]{\smash{0.4}}}%
    \put(0.52334404,0.35336766){\makebox(0,0)[lb]{\smash{0.6}}}%
    \put(0.52354327,0.42885361){\makebox(0,0)[lb]{\smash{0.8}}}%
    \put(0.52540279,0.50433955){\makebox(0,0)[lb]{\smash{1.0}}}%
    \put(0.30338152,0.00659342){\color[rgb]{0,0,0}\makebox(0,0)[b]{\smash{(a) Real data}}}%
    \put(0.77889953,0.00659342){\color[rgb]{0,0,0}\makebox(0,0)[b]{\smash{(b) Synthetic data}}}%
    \put(0.09783192,0.54661589){\color[rgb]{0,0,0}\makebox(0,0)[lb]{\smash{PPF}}}%
    \put(0.46117486,0.54661589){\color[rgb]{0,0,0}\makebox(0,0)[lb]{\smash{LINEMOD+}}}%
    \put(0.26323072,0.54661589){\color[rgb]{0,0,0}\makebox(0,0)[lb]{\smash{PPF PP}}}%
    \put(0.75092623,0.54661589){\color[rgb]{0,0,0}\makebox(0,0)[lb]{\smash{LINEMOD+ PP}}}%
  \end{picture}%
\endgroup%

\caption{\label{fig:perfs_real_vs_synthetic}Comparison of performances obtained with real and synthetic data. Precision-recall curves for the retrieval of less than 50\% occluded instances.}
\end{figure}

Because we used simple and ideal material and lighting in our CGI renderings, the synthetic data produced is different from the real one, and of slightly better quality (96.6\% of the pixels with depth information, against 96.2\% for the real data). This quantitatively affects performances and every pose estimation method we tested showed slightly better performances on the synthetic dataset than on the real one (see \emph{markers flat} results in table~\ref{tab:performances}).
Nonetheless, synthetic data remains plausible thanks to our depth sensor simulation procedure, and the performances of the evaluated methods compare to one another very similarly on both datasets, as illustrated figure~\ref{fig:perfs_real_vs_synthetic}. We therefore consider the synthetic data produced to be \emph{realistic enough} for comparative evaluation purposes.


\begin{table*}
\caption{\label{tab:performances} Performances of two object detection and pose estimation methods, and their variants exploiting object symmetries (\emph{sym}) if any, and post-processing of the best 20 pose hypotheses. Accounting for symmetries improves performances of both methods.}
\vspace{-5mm}
\centering\footnotesize
\setlength\tabcolsep{2.0pt} 
\renewcommand{\arraystretch}{1.2}
\begin{tabular}{|c|c|ccc|ccc||ccc|ccc|||ccc|ccc||ccc|ccc|}
\cline{3-26}
\multicolumn{1}{c}{}      & \multicolumn{1}{c|}{}     & \multicolumn{12}{c|||}{Raw method}                                                                                                                                                                                                                                                                                                               & \multicolumn{12}{c|}{With post-processing of the best 20 pose hypotheses (PP)} \\
\multicolumn{1}{c}{}      & \multicolumn{1}{c|}{}     & \multicolumn{3}{c|}{PPF}                                                           & \multicolumn{3}{c||}{PPF \emph{sym}}                                                       & \multicolumn{3}{c|}{LINEMOD+}                                                         & \multicolumn{3}{c|||}{LINEMOD+ \emph{sym}}                                                     & \multicolumn{3}{c|}{PPF}                                                           & \multicolumn{3}{c||}{PPF \emph{sym}}                                                       & \multicolumn{3}{c|}{LINEMOD+}                                                         & \multicolumn{3}{c|}{LINEMOD+ \emph{sym}} \\
\cline{2-2}
\multicolumn{1}{c|}{}      & Dataset                   & AP                        & AP1                       & AP3                       & AP                        & AP1                       & AP3                       & AP                        & AP1                       & AP3                       & AP                        & AP1                       & AP3                       & AP                        & AP1                       & AP3                       & AP                        & AP1                       & AP3                       & AP                        & AP1                       & AP3                       & AP                        & AP1                       & AP3 \\
\hline
\multirow{5}[0]{*}{\begin{sideways}Real data\end{sideways}} & markers bump              & .35                       & .66                       & .43                       & \textbf{--}               & \textbf{--}               & \textbf{--}               & .85                       & 1.00                      & .96                       & \textbf{--}               & \textbf{--}               & \textbf{--}               & .56                       & .97                       & .84                       & \textbf{--}               & \textbf{--}               & \textbf{--}               & .91                       & 1.00                      & .99                       & \textbf{--}               & \textbf{--}               & \textbf{--} \\
                          & markers clutter           & .34                       & .36                       & .31                       & \textbf{--}               & \textbf{--}               & \textbf{--}               & .57                       & .67                       & .53                       & \textbf{--}               & \textbf{--}               & \textbf{--}               & .52                       & .70                       & .52                       & \textbf{--}               & \textbf{--}               & \textbf{--}               & .68                       & .83                       & .69                       & \textbf{--}               & \textbf{--}               & \textbf{--} \\
                          & markers flat              & .26                       & .54                       & .31                       & \textbf{--}               & \textbf{--}               & \textbf{--}               & .83                       & .99                       & .97                       & \textbf{--}               & \textbf{--}               & \textbf{--}               & .46                       & .94                       & .76                       & \textbf{--}               & \textbf{--}               & \textbf{--}               & .90                       & 1.00                      & .99                       & \textbf{--}               & \textbf{--}               & \textbf{--} \\
                          & juice~\citep{doumanoglou2016}                    & .04                       & .15                       & .07                       & \textbf{--}               & \textbf{--}               & \textbf{--}               & .01                       & .01                       & .01                       & \textbf{--}               & \textbf{--}               & \textbf{--}               & .07                       & .29                       & .11                       & \textbf{--}               & \textbf{--}               & \textbf{--}               & .06                       & .24                       & .10                       & \textbf{--}               & \textbf{--}               & \textbf{--} \\
                          & coffee cup~\citep{doumanoglou2016}               & .16                       & .76                       & .53                       & \textbf{.28}              & \textbf{.96}              & \textbf{.85}              & .03                       & .37                       & .10                       & \textbf{.08}              & .37                       & \textbf{.17}              & .23                       & .98                       & .90                       & \textbf{.30}              & \textbf{1.00}             & \textbf{.92}              & .10                       & .95                       & .61                       & \textbf{.20}              & \textbf{1.00}             & \textbf{.93} \\
\hline
\multirow{9}[0]{*}{\begin{sideways}Synthetic data\end{sideways}} & markers flat              & .29                       & .55                       & .36                       & \textbf{--}               & \textbf{--}               & \textbf{--}               & .87                       & .99                       & .97                       & \textbf{--}               & \textbf{--}               & \textbf{--}               & .50                       & .94                       & .79                       & \textbf{--}               & \textbf{--}               & \textbf{--}               & .91                       & .99                       & .99                       & \textbf{--}               & \textbf{--}               & \textbf{--} \\
                          & tless 22                  & .08                       & .52                       & .34                       & \textbf{--}               & \textbf{--}               & \textbf{--}               & .19                       & .63                       & .54                       & \textbf{--}               & \textbf{--}               & \textbf{--}               & .12                       & .89                       & .76                       & \textbf{--}               & \textbf{--}               & \textbf{--}               & .21                       & .81                       & .81                       & \textbf{--}               & \textbf{--}               & \textbf{--} \\
                          & bunny                     & .29                       & .83                       & .66                       & \textbf{--}               & \textbf{--}               & \textbf{--}               & .39                       & .97                       & .94                       & \textbf{--}               & \textbf{--}               & \textbf{--}               & .37                       & .99                       & .97                       & \textbf{--}               & \textbf{--}               & \textbf{--}               & .45                       & .99                       & .98                       & \textbf{--}               & \textbf{--}               & \textbf{--} \\
                          & tless 20                  & .10                       & .49                       & .35                       & \textbf{.20}              & \textbf{.82}              & \textbf{.64}              & .17                       & .81                       & .44                       & \textbf{.25}              & .81                       & \textbf{.75}              & .14                       & .92                       & .84                       & \textbf{.23}              & \textbf{.98}              & \textbf{.94}              & .24                       & 1.00                      & .97                       & \textbf{.31}              & 1.00                      & \textbf{.99} \\
                          & tless 29                  & .15                       & .69                       & .40                       & \textbf{.19}              & \textbf{.76}              & \textbf{.56}              & .14                       & .71                       & .34                       & \textbf{.20}              & .71                       & \textbf{.50}              & .21                       & .90                       & .76                       & \textbf{.23}              & \textbf{.91}              & \textbf{.79}              & .20                       & .88                       & .84                       & \textbf{.26}              & \textbf{.92}              & \textbf{.86} \\
                          & brick                     & .05                       & .24                       & .13                       & \textbf{.08}              & \textbf{.35}              & \textbf{.22}              & .20                       & .97                       & .47                       & \textbf{.31}              & .97                       & \textbf{.76}              & .10                       & .68                       & .47                       & \textbf{.13}              & \textbf{.77}              & \textbf{.59}              & .32                       & .98                       & .95                       & \textbf{.39}              & \textbf{.99}              & \textbf{.96} \\
                          & gear                      & .24                       & .42                       & .30                       & \textbf{.62}              & \textbf{.94}              & \textbf{.89}              & .15                       & .93                       & .31                       & \textbf{.44}              & \textbf{.95}              & \textbf{.84}              & .30                       & .81                       & .76                       & \textbf{.63}              & \textbf{.99}              & \textbf{.97}              & .25                       & .99                       & .92                       & \textbf{.50}              & .99                       & \textbf{.98} \\
                          & candlestick               & .09                       & .32                       & .22                       & \textbf{.16}              & \textbf{.60}              & \textbf{.47}              & .17                       & .86                       & .29                       & \textbf{.38}              & \textbf{.92}              & \textbf{.78}              & .15                       & .85                       & .75                       & \textbf{.22}              & \textbf{.85}              & \textbf{.78}              & .26                       & 1.00                      & .96                       & \textbf{.49}              & 1.00                      & \textbf{1.00} \\
                          & pepper                    & .04                       & .08                       & .06                       & \textbf{.06}              & \textbf{.25}              & \textbf{.13}              & .03                       & .11                       & .05                       & \textbf{.04}              & \textbf{.11}              & \textbf{.08}              & .08                       & .68                       & .38                       & \textbf{.12}              & \textbf{.85}              & \textbf{.57}              & .03                       & .13                       & .07                       & \textbf{.03}              & \textbf{.14}              & \textbf{.08} \\
\hline
\end{tabular}%
\\
AP: Average Precision for the retrieval of instances less than 50\% occluded.\\
AP$n$ ($n \in \mathbb{N}^*$): Average Precision given at most $n$ results returned for the retrieval of instances less than 50\% occluded. 
\end{table*}

\subsection{Effect of accounting for object symmetries}
\label{subsec:results}

While precision-recall curves such as the ones figure~\ref{fig:perfs_real_vs_synthetic} provide a finer-grained understanding of performances, we synthesize our results for the sake of readability in table~\ref{tab:performances} through different metrics. We consider the \emph{Average Precision} (AP)~\citep{average_precision}, a usual metric consisting in the area under the precision-recall curve, given the goal of retrieval of instances less than 50\% occluded.
One might only be interested in the retrieval of a few object instances -- \eg for robotics manipulation -- therefore we also present the Average Precision given at most $n \in \{1, 3 \}$ results returned (AP$n$). The formalism enabling to compute those metrics is defined in section~\ref{sec:evaluation_methodology}.

\paragraph{Results discussion}
As expected, adding a basic post-processing step (PP) to both PPF and LINEMOD+ significantly improves performances on every dataset (with an average AP improvement of respectively +59\% and +60\%).
Taking object symmetries into consideration also leads to great improvements for the PPF method, with or without postprocessing and regardless of the considered metrics (\eg respectively +18\% and +99\% increases of AP3).
The PPF method indeed relies on the aggregation of multiple weak pose hypotheses in order to generate stronger ones, and considering symmetries significantly helps in this aggregation, as discussed in section~\ref{subsec:considering_symmetries_for_pose_estimation}.

LINEMOD+ also benefits from considering symmetries (+54\% and +93\% increases of AP respectively with and without post-processing). Since symmetry considerations are only used to filter out duplicates prior to pose refinement, we should not observe performance improvements when retrieving at most one pose hypothesis without post-processing for this method. The small difference observed here (+1\% increase of AP1) is merely an artifact of our implementation, which includes a basic filtering step of pose hypotheses outside the frustum of the camera after pose refinement.
Accounting for symmetries however benefits in the other cases to both precision, by removing duplicates, and recall, as it allows to consider more truly different poses for a given number of pose hypotheses, a point we observe in the global performance improvements of LINEMOD+ PP (+54\% AP and +11\% AP3) which already includes a duplicates filtering step, even without symmetry considerations.

\section{Conclusion}

We focused in this article on the evaluation of 3D rigid object detection and pose estimation techniques, in scenes containing an arbitrary number of instances, in arbitrary poses. We proposed two methods to generate automatically annotated datasets of such scenes, and metrics suited for the performance evaluation of this generic scenario, even in the case of symmetric objects. We showed how those symmetry considerations could be adapted within existing pose estimation methods themselves, and our experimental results suggest that it leads to significant performance improvements.

{\small
\bibliographystyle{MyIEEEtranN} 
\bibliography{biblio}
}

\end{document}